\documentclass[10pt,twocolumn,letterpaper]{article}

\usepackage{cvpr}              % To produce the CAMERA-READY version
\usepackage[dvipsnames]{xcolor}

\definecolor{cvprblue}{rgb}{0.21,0.49,0.74}
\usepackage[pagebackref,breaklinks,colorlinks,citecolor=cvprblue]{hyperref}

\usepackage{algorithm}
\usepackage{algorithmic}
\usepackage{multirow}
\usepackage{amssymb}
\usepackage{subcaption}

\newcommand{\reid}{re-id}
\newcommand{\sota}{the state-of-the-art}
\newcommand{\myparagraph}[1]{\noindent\textbf{#1.}}
\newcommand{\name}{{DeepChange}}
\def\ie{i.e.}
\def\eg{e.g.}

\usepackage{bm}

\def\f{\bm{f}}
\def\v{\bm{v}}
\def\p{\bm{p}}
\def\q{\bm{q}}

\def\x{\bm{x}}

\def\z{\bm{z}}

\def\u{\bm{u}}

\def\II{\mathcal{I}}
\def\I{\mathbf{I}}

\def\RR{\mathbb{R}}
\def\A{\mathcal{A}}

\def\C{\mathcal{C}}
\def\D{\mathcal{D}}

\def\M{\mathcal{M}}
\def\N{\mathcal{N}}

\def\S{\mathcal{S}}
\def\X{\mathcal{X}}
\def\Y{\mathcal{Y}}

\def\paperID{*****} % *** Enter the Paper ID here
\def\confName{CVPR}
\def\confYear{2024}

\title{SiCL: Silhouette-Driven Contrastive Learning for Unsupervised Person Re-Identification with Clothes Change}

\author{
    Mingkun Li$^{\dag,1}$, Peng Xu$^{*,2}$, Chun-Guang Li$^{\dag,1}$, Jun Guo $^{\dag,1}$\\
    $^{1}$Beijing University of Posts and Telecommunications,
    $^{2}$Tsinghua University\\
    {\tt\small $^\dag$\{mingkun.li,lichunguang, guojun\}@bupt.edu.cn,$^*$peng$\_$xu@tsinghua.edu.cn}
}

\begin{document}
\maketitle
\begin{abstract}
In this paper, we address a highly challenging yet critical task: unsupervised long-term person re-identification with clothes change.
Existing unsupervised person re-id methods are mainly designed for short-term scenarios and usually rely on RGB cues so that fail to perceive feature patterns that are independent of the clothes.
To crack this bottleneck, we propose a silhouette-driven contrastive learning (SiCL) method, 
which is designed to learn cross-clothes invariance by integrating both the RGB cues and the silhouette information within a contrastive learning framework.
To our knowledge, this is the first tailor-made framework for unsupervised long-term clothes change \reid{}, with superior performance on six benchmark datasets.  We conduct extensive experiments to evaluate our proposed SiCL compared to the state-of-the-art unsupervised person \reid{} methods across all the representative datasets. Experimental results demonstrate that our proposed SiCL significantly outperforms other unsupervised re-id methods. 
\end{abstract}    
\section{Introduction}
\label{sec:intro}

Person re-identification (\reid{}) aims to match  person identities of bounding box images that are captured from distributed camera views~\cite{ye:surveyarixv2020}. 
Most conventional studies 
%in the field of 
for unsupervised person re-id have only focused on the scenarios {\em without} clothes change~\cite{zhao:PAMI16,ge2018fd}.
However, re-id in such scenarios is unrealistic since 
% the majority of 
people may often change their clothing.
% , if not more often. 
Thus, these studies for the scenarios {\em without} clothes change may only be useful in the short-term re-id settings but fail 
% when facing 
to handle
the long-term person \reid{} for which we have to face the scenarios with clothes changes. 

Recently, there have been %recent 
some attempts to address the long-term \reid{} task~\cite{Change:1,change:8}. However, all of these attempts employ supervised learning methods which heavily rely on large amounts of labelled training data.
Unfortunately, collecting and annotating person identity labels under the scenario of unconstrained clothes change is extremely difficult. Preparing labelled re-id training data, \eg, \name{}~\cite{Deepchange}, in a realistic scenario is quite expensive and exhausting.

\begin{figure}[t]
\centering
\includegraphics[width=0.99\linewidth]{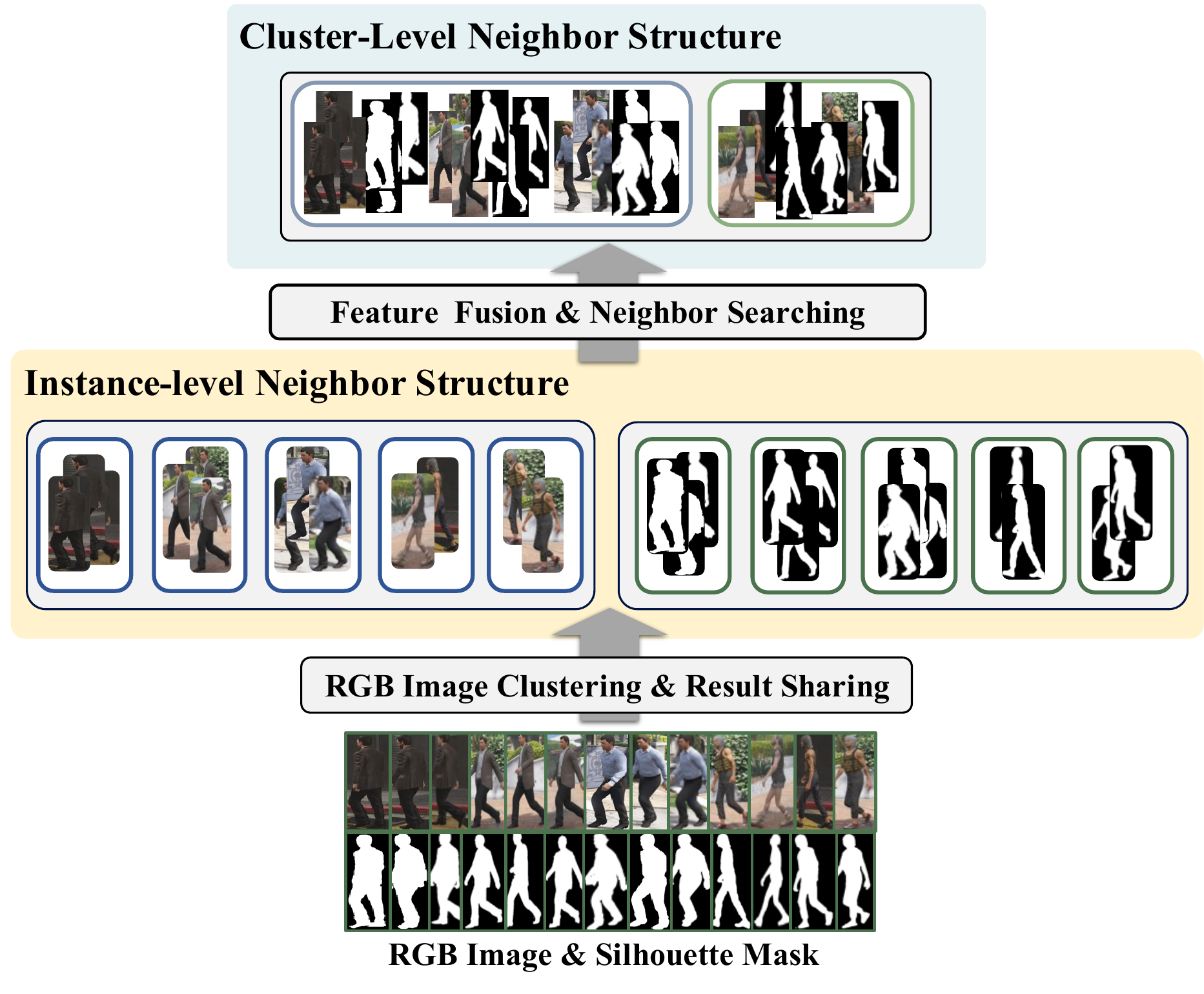}
\caption{Illustration of hierarchical fusion clustering: At a lower level, clustering is used to reveal the neighbor structure of instances. At a higher level, silhouette features and RGB features are integrated %combined 
to exploit %expose 
the neighbor structure of the clusters.}
\label{fig:semantic_test}
\end{figure}

\begin{figure}
\centering
{\includegraphics[width=0.99\linewidth]{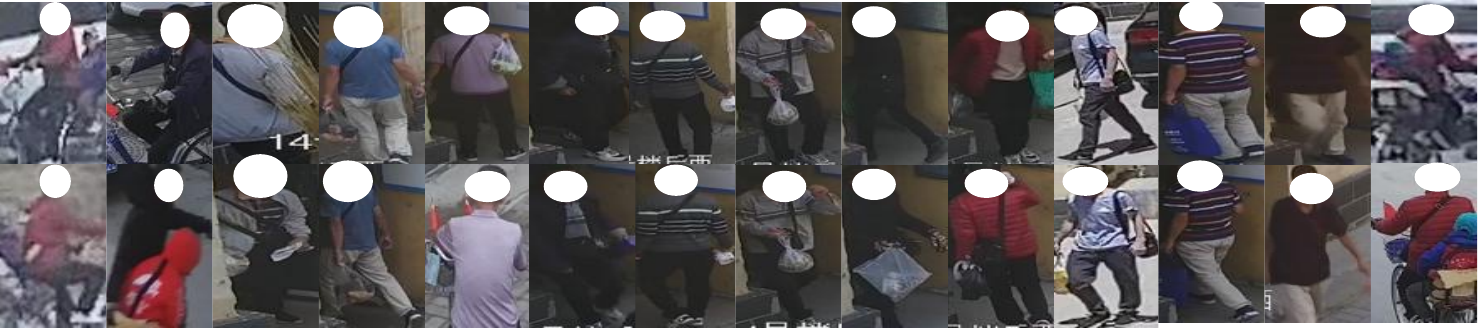}}\\
\caption{Visualizing the intrinsic challenges in long-term person \reid{} with clothes change. 
We randomly selected $28$ images of a single individual from the \name{} \cite{Deepchange} dataset.
Evidently, the variances in visual characteristics between different outfits (rows) are considerably more pronounced compared to those within the same outfit (each column).
}
\label{fig:clothes-change}
\end{figure}

Due to the significance of long-term person \reid{}, 
it is appealing to develop unsupervised method to address %approach 
long term person re-id problem %that eliminates 
without the tedious requirement of person identity labeling. 
This is a more complex but more realistic extension of the previous unsupervised short-term person re-id~\cite{li2018unsupervised,Ge:NIPS20, lmkpr} that different people may have similar clothes whilst the same person might wear variant clothes with very distinct appearances, as shown in Fig.~\ref{fig:clothes-change}.
Unfortunately, prior investigations on unsupervised person re-identification have %regrettably 
neglected situations involving clothing changes. Conventional approaches are incapable of capturing clothing-independent patterns since they solely rely on RGB features as their driving force~\cite{TIPlmk}. 
Specifically, the majority of existing unsupervised methods~\cite{dai:ACCV22} are cluster-oriented and tend to yield feature extractions primarily dominated by color~\cite{TIPlmk}.

As a consequence, the clustering algorithm will blindly assign every training sample with a color-based pseudo label, which is error-prone with a large cumulative propagation risk and ultimately leads to sub-optimal solutions, such as simply grouping the individuals who wear the same clothing.

In this paper, we propose a novel silhouette-driven contrastive learning framework, termed SiCL, for attacking the challenging task of unsupervised person re-id in long-term setting.  
In SiCL, we incorporate the person silhouette and RGB images into a contrastive learning framework to learn cross-clothes invariance features. 
Specifically, SiCL employs a dual-branch network to perceive both silhouette and RGB image features and incorporates them %both features 
to construct a hierarchical neighbor structure, as demonstrated in Fig.~\ref{fig:semantic_test}. The RGB feature is used to construct the low-level instance neighbor structure, and the fused features are used to construct the high-level cluster neighbor structure.
Different from other methods that are only relying on RGB patterns to construct the single-level neighbor structure, in SiCL, %our feature fusion operation 
we incorporates clothing-independent features hidden in the silhouette. Such an %This 
additional feature can %aids in providing 
provide further guidance to %the model in the learning of 
modeling the invariant features across different clothes.
Moreover, we introduce a contrast-learning module to learn invariant features between the silhouette and RGB images at various neighbor structure levels.

To validate the effectiveness of our SiCL approach, we conduct comprehensive evaluation on six long-term person re-id datasets comparing to the state-of-the-art unsupervised person re-identification (re-id) methods. Experimental results demonstrate the superior performance of our approach. Notably, our SiCL can not only outperforms all short-term methods significantly, but also achieves comparable performance to the \sota~fully supervised methods.

The main contribution of the paper can be summarized as follows.

\begin{enumerate}

\item We propose a silhouette-driven contrastive learning framework for unsupervised long-term person \reid{} with clothes change. To the best of our knowledge, this is the first work to investigate unsupervised long-term person \reid{} with clothes change.

\item We propose to incorporates both person silhouette information and hierarchical neighbor structure into a contrastive learning framework to guide the model for learning cross-clothes invariance feature.

\item We conduct extensive experiments on six representative datasets to evaluate the performance of the proposed SiCL and the state-of-the-art unsupervised re-id methods. 

\end{enumerate}

\section{Related Work}
\label{sec:related-work}

\subsection{Long-Term Person Re-Identification}
A few recent person \reid{} studies attempt  to tackle~\cite{Change:1,change:7} the long-term clothes-changing situations via supervised training, and emphasize the use of additional supervision for the general appearance features (\eg, clothes, color), to  enable  the model to learn the cross-clothes features. For example, 
\cite{Change:1} generates contour sketch images from RGB images and highlights the invariance between the sketch and RGB.
\cite{change:8} explores fine-grained body shape features by estimating masks with discriminative shape details and extracting pose-specific features. 
While these seminal works provide inspiring attempts for \reid{} with clothes change, there are still some limitations: 
a) Generating contour sketches or other side-information requires additional model components and will cause extra computational cost; 
b) To the best of our knowledge, all models designed for \reid{} clothes change are supervised learning methods, 
and thus lack of generalization ability to the open-world dataset settings. 
In this paper, we attempt to tackle the challenging \reid{} with clothes change under unsupervised setting.

\subsection{Unsupervised Person Re-Identification}
To avoid the high consumption in data labeling,
a large and growing body of literature has investigated unsupervised person re-id~\cite{Zheng:arXiv16,Wang:CVPR18}.  
The existing unsupervised person Re-ID methods can be divided into two categories: a) unsupervised domain adaptation methods, which require a labeled source dataset and an unlabelled target dataset~\cite{Peng:CVPR16,Wang:CVPR18}; and b) purely unsupervised methods that work with only an unlabelled dataset~\cite{PPLR,dai:ACCV22}. 
However, up to date, 
the unsupervised person re-id methods are focusing on short-term scenarios, none of them taking into account the long-term re-id scenario. 
To the best of our knowledge, this is the first attempt to address the long-term person re-id in unsupervised setting. Thus, as a byproduct, we systematically and comprehensively evaluate the performance of the existing \sota{} unsupervised methods~\cite{Ge:NIPS20,PPLR} on six long-term person re-id datasets 
and set up a preliminary benchmarking ecosystem for the long-term person \reid{} community.

\begin{figure*}[tb]
\centering
 \includegraphics[width=0.99\linewidth]{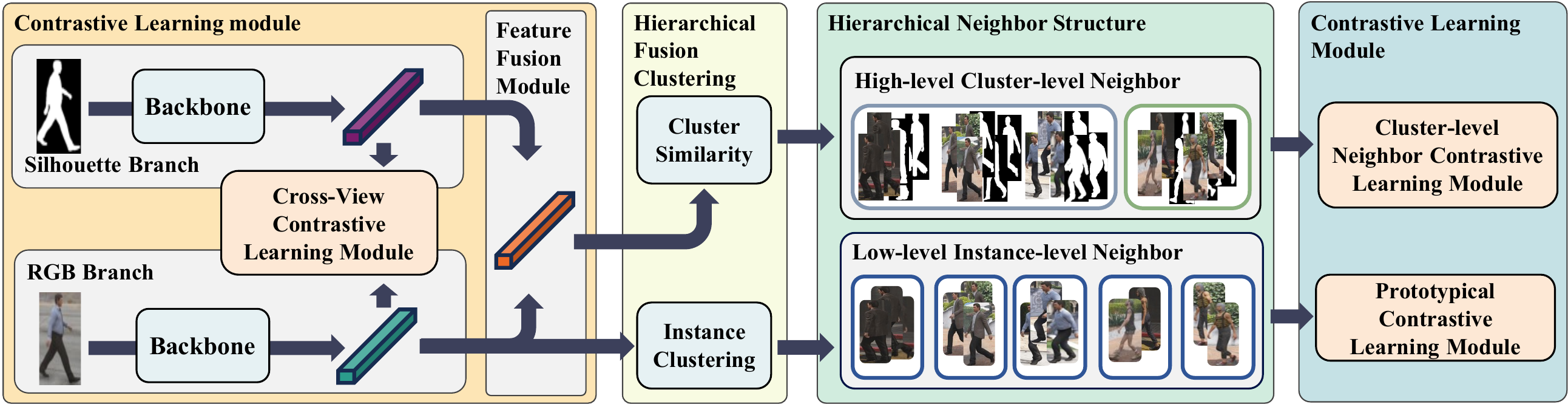}
   \caption{
   Our proposed  Silhouette-Driven Contrastive Learning (SiCL) framework.
   }
\label{fig:framework}
\end{figure*}
% methodology
\section{Our Methodology: Silhouette-Driven Contrastive Learning (SiCL)}

{ Our SiCL approach integrates silhouette and RGB images into a contrastive learning framework assisted with the guidance from hierarchical clustering structure.
}
For clarity, we provide a flowchart of our SiCL in Fig.~\ref{fig:framework}. Our SiCL consists of two network branches: $F(\cdot|\Theta)$ and $F^\prime(\cdot|\Theta^\prime)$, which perceive RGB and 
silhouette patterns, respectively.
We use $\Theta$ and $\Theta^\prime$ to denote the corresponding network parameters. 
In our SiCL, the parameters $\Theta$ and $\Theta^\prime$ are learned separately (without parameters sharing).  We design a predictor layer $G(\cdot|\Psi)$ which follows the RGB branch, where $\Psi$ denotes the parameters in the predictor layer.
In addition, we also design a feature fusion layer $R(\cdot|\Omega)$, where $\Omega$ denotes the parameters.

Given an unlabeled pedestrian image dataset $\II = {\{I_i}\}_{i=1}^N$ consisting of $N$ samples,  we generate the corresponding pedestrian silhouette mask dataset $\S = {\{S_i}\}_{i=1}^N$  through human parsing network such as SCHP~\cite{ParsingNet:TPAMI20}.  
For an input image $I_i\in \II$,  we use the $I_i$ as the input of $F(\cdot|\Theta)$ and $S_i$ as the inputs of $F^\prime(\cdot|\Theta^\prime)$.
For simplicity, we denote the output features of the  $F(\cdot|\Theta)$ and $F^\prime(\cdot|\Theta^\prime)$ as $\x_i$ and $ \bm{\tilde x}_i$, 
denote the output of the predictor layer in the $G(\cdot|\Psi)$ as $\z_i$, and denote the output of fusion layer $R(\cdot|\Omega)$ as $\bm{f}_i$, where $\x_i, \bm{\tilde x}_i, \z_i, \f_i \in R^D$.
% {respectively}. 

% \red
As illustrated in Fig. \ref{fig:framework}, our SiCL compose of a contrastive learning module and a hierarchical fusion clustering module.  
In hierarchical fusion clustering module, the hierarchical neighbor structure is generated by leveraging RGB and silhouette features at both the instance-level and cluster-level.
In contrastive learning module, we introduce the person's silhouette and use the hierarchical neighbor structure as supervision information to train model, and investigate cross-clothes features via contrastive learning.
Furthermore, we build an adaptive learning strategy to automatically modify the hierarchical neighbor selection, which can flexibly select neighbor clusters according to dynamic criteria.

To store the outputs of the two branches and the fusion layer, in SiCL, we maintain the three instance-memory banks $\M = \{ {\v}_i\}_{i=1}^N$, $\tilde \M = \{  \bm{\tilde v}_i \}_{i=1}^N$, and $\hat \M = \{  \bm{\hat v}_i\}_{i=1}^N$, respectively, where $ {\v}_i,   \bm{\tilde v}_i, \bm{\hat v}_i \in R^D$. 
Memory banks $\M$ and $\hat \M$ are initialized with $\X:=\{\x_1,\cdots,\x_N\}$ and $\tilde \M$ is %are
initialized with $\tilde \X:=\{\bm{\tilde x}_1,\cdots,\bm{\tilde x}_N\}$, where $\X$ and $\tilde \X$ are the outputs of  $F(\cdot|\Theta)$ and $F^\prime(\cdot|\Theta^\prime)$.

\subsection{Hierarchical Fusion Clustering}
\label{sec: Hierarchical Semantic}

To construct a hierarchical neighbor structure, we sort the hierarchical neighbor structure into two levels: a) low-level instance neighbors, and b) high-level cluster neighbors.

\myparagraph{Low-level Instance Neighbors} We use 
$F(\cdot|\Theta)$ outputs 
to 
construct
$m$ clusters
$\C:=\{\C^{(1)}, \C^{(2)}, \cdots, \C^{(m)}\}$.
The clustering result $\C$ indicates the connection between neighbors at the instance level since that if samples are clustered together, it also indicates that their RGB features are similar. In the subsequent training process, this clustering result $\C$ will also be used to guide the training of  branch $F^\prime(\cdot|\Theta^\prime)$. 

\myparagraph{High-level Cluster Neighbors} 
Since that the silhouette masks contain richer clothing invariance features, we use them to find those pedestrian samples that are similar at the cluster level, \eg, the same person wearing different clothing. To be specific, we fuse the RGB feature and the silhouette feature to renew the representation of instance, and search the cluster neighbor based on fusion features. 
For an image $\I_i$, the fusion feature $f_i$ is defined as:
\begin{equation}
f_i = F(concate(\x_i|\bm{\tilde x}_i)),
\label{eq:feature-fusion}
\end{equation}
where $concate(\cdot)$ denotes the channel-wise concatenating
$\x_i$ and $\bm{\tilde x}_i$.
Based on the fused features, we define the cluster center ${\u}_{\omega(\I_i)}$ 
as
\begin{equation}
{\u}_{\omega(\I_i)} = \frac{1}{|\C^{(\omega(\I_i))}|} \sum_{I_j \in \C^{(\omega(\I_i))}}\f_j,
\label{eq:cluster_center_fusion}
\end{equation}
where $\omega(\I_i)$ is the cluster index of image $\I_i$. 

Once constructing the cluster centre ${U}= {\{{\u_i}}\}_{i=1}^m$, we will find the cluster-level nearest neighbours based on the similarity of the cluster centres 
Once contrasting the cluster-center , we find cluster-level  neighbors and construct the neighbor set $\N$. 
Specifically, for cluster $\C^{(\ell)}$, we define %metric 
the similarity %score 
between clusters $\C^{(\ell)}$ and $\C^{(i)}$ as 
\begin{equation}
\D(\C^{(\ell)}, \C^{(i)}) = \frac{{\u}_\ell^{\top}}{\|{\u}_{\ell}\|_2} \frac{ {\u}_i}{\| {\u}_i \|_2}.
\label{eq:cluster_sim}
\end{equation}
We denote the cluster neighbor set of cluster $\C^{(\ell)}$ as $\N^{(\ell)}$, 
which includes the top-$k$ similar clusters $\C^{(i)}$ sorted by $\D(\C^{(\ell)}, \C^{(i)})$. 
Then, we form the total cluster-level neighbor set
$\A:= \{ \N^{(1)},\cdots, \N^{(m)} \}$. 
This process is known as hierarchical fusion clustering due to the use of a fused feature operation and the construction of a multi-level neighbor structure.

In the hierarchical fusion clustering stage, we construct  neighbor structures based on some specific clustering algorithm and cluster-level neighbor searching, respectively. The clustering result of the output features $\X:=\{\x_1,\cdots,\x_N\}$ from $F(\cdot|\Theta)$ is used to generate the pseudo labels $\Y :=\{y_1,\cdots,y_N\}$, and cluster-level neighbor set $\A$ contains the neighbor index for each cluster. 
In the contrastive learning stage, the hierarchical neighbor structure is used as supervision information to guide the model learn cross-clothes features.

\subsection{Contrastive Learning Module }
\label{sec:Contrastive-Learning-Framework}

To effectively explore the invariance features between RGB images and silhouette masks, we construct three contrastive learning modules to train SiCL assisted with the self-supervision information provided by the hierarchical fusion clustering as follows:
a) Prototypical contrastive learning module, which is used for contrast training between positive samples and negative pairs;
b) Cross-view contrastive learning module, which is used for contrast training between RGB images and silhouette masks; and
c) Cluster-level neighbor contrastive learning module, which is used for contrast training between cluster-level neighbor clusters and negative pairs.

% use the prototypical 
\myparagraph{Prototypical Contrastive Learning Module}  We apply prototypical contrastive learning to discover the hidden information inside the cluster structure. For the $i$-th instance, we denote its cluster index as $\omega(I_i)$, the center of $C^{\omega(I_i)}$ as the positive prototype, and all other cluster centers as the negative prototypes. 
We define the prototypical contrastive learning loss as follows:
\begin{equation}
 \mathcal{L}_P =  - \sum_{\q^* \in {\{\q, \bm{\tilde q}, \bm{\hat q\}}}} (1 - \q^*_i)^2 \ln( \q^*_i), 
\label{eq:prototype-loss}
\end{equation}
where $\q_{i}$, $\bm{\tilde q}_i$ and $\bm{\hat q}_i$ measure the consistency between the outputs of $F(\cdot|\Theta)$, $F^\prime(\cdot|\Theta^\prime)$ and $R(\cdot|\Omega)$ and the related prototype computed with the memory bank and are defined as 
\begin{equation}
 \q_{i}=\frac{\exp(\p_{{\omega(I_i)}}^\top{\x_i}/\tau)}{ \sum_{\ell = 1}^{m}\exp(\p_{\ell}^\top {\x_i}/\tau)},
\label{eq:softmax-cal}
\end{equation}
where $\p_{{\omega(I_i)}}$ as the RGB prototype vector of the cluster $\C^{(\omega(I_i))}$ is defined by
\begin{equation}
\p_{\omega(I_i)} = \frac{1}{|\C^{(\omega(I_i))}|} \sum_{I_j \in \C^{(\omega(I_i))}} {\v}_j,
\label{eq:cluster_center}
\end{equation}
here ${\v}_j$ is the instance feature of image $I_j$ in $\M$, $\bm{\tilde p}_i$ and $\bm{\hat p}_i$ are calculated in the same way with corresponding instance memory bank $\tilde \M$ and $\hat \M$, respectively.

The prototypical contrastive learning module performs 
contrastive learning between positive and negative prototypes to improve the discriminant ability for the networks $F(\cdot|\Theta)$ and $F^\prime(\cdot|\Theta^\prime)$ and the feature fusion layer $R(\cdot|\Omega)$.

\myparagraph{Cross-view Contrastive Learning Module} To effectively train the contrastive learning framework across the two views, we design a cross-view contrastive module to mine the invariance between RGB images and silhouette masks. To match the feature outputs of two network branches at both the instance level and cluster level, specifically, 
we introduce the negative cosine similarity of the outputs of $G(\cdot|\Psi)$ and $F^\prime(\cdot|\Theta^\prime)$ to define the two-level contrastive losses as follows:
\begin{equation}
\mathcal{L}_C := -  \frac{\z^\top_i}{\|\z_i\|_2} \frac{\bm{\tilde x}_i}{\| \bm{\tilde x}_i \|_2} 
 - \frac{\z_i^\top}{\|\z_i\|_2} \frac{\bm{\tilde p}_{\omega(I_i)}}{\| \bm{\tilde p}_{\omega(I_i)} \|_2}, % \big)
\label{eq:contrastive-loss}
\end{equation}
where $\| \cdot \|_2$ is the $\ell_2$-norm.

The cross-view contrastive learning module 
explores the invariance between RGB images and silhouette masks and thus %This will 
assist the network to mine the invariance information provided by the RGB image 
and the 
silhouette mask, as well as imposing such a self-supervision %this supervisory 
information on the module for learning clothing-unrelated features.

\myparagraph{Cluster-level Neighbor Contrastive Learning Module}
To avoid training the model into degeneration 
that only push samples with similar appearance together, we design a cluster-level neighbor contrastive learning module.
Particularly, we propose a weighted cluster-level neighbor contrastive loss 
as follows: 
\begin{equation}
\mathcal{L}_N = -\sum_{j \in \N^{(\omega(I_i))}} w_{ij} (\frac{\z_i^\top}{\|\z_i\|_2} \frac{\bm{\tilde p}_{j}}{\| \bm{\tilde p}_{j} \|_2)}), 
\label{eq:Neighour_loss}
\end{equation}
where $\N^{(\omega(I_i))}$ is the set of neighbors of cluster $\omega(I_i)$, and $w_{ij}$ is the weight, which is defined as
\begin{equation}
w_{ij} = \D(\C^{(i)},\C^{(j)}),
\label{eq:Neighour_loss_detail}
\end{equation}
in which $ \D(\C^{(i)},\C^{(j)})$ is defined in Eq.~\ref{eq:cluster_sim}.
Owing to the Cluster-level neighbor contrastive learning module trained via
loss in Eq.~\ref{eq:Neighour_loss}, the cluster-level neighbors will be pushed closer in the feature space. This will help the model to investigate consistency across different neighbor clusters.

\subsection{Curriculum Nerighour Selecting }
\label{sec:Curriculumneighbor}
Since in the early training stages, the model has a weaker ability to distinguish samples, we hope the ability improves during training. 
To this end, we provide a curriculum strategy for searching neighbors which sets the search range according to the training progress. 
Specifically, we set the cluster-level neighbor searching range $k$ as
\begin{equation}
k:=  t  \lfloor K / T \rfloor, 
\label{eq:Curriculumneighbor}
\end{equation}
where $T$ is the total number of training epochs and $t$ is the current step, $K$ is a hyper-parameter.

\subsection{Training and Inference Procedure for SiCL}

\myparagraph{Training Procudure}
In SiCL, the two branches are implemented with ResNet-50 \cite{he:CVPR16resnet} and do not share the parameters. We first pre-train the two network branches on ImageNet and use the learned features to initialize the three memory banks $\M$, $\tilde \M$, and $\hat \M$, respectively. 
In the training phase, we train both network branches and the fusion layer with loss:
\begin{equation}
\mathcal{L} := \mathcal{L}_{P} + \mathcal{L}_{C} + \mathcal{L}_{N}. 
\label{eq:AllLoss}
\end{equation}
We update the three instance memory banks $\M$, $\tilde \M$ and $\hat \M$, respectively, as follows:
\begin{equation}
{\v}_i^{(t)} \leftarrow \alpha {\v}_i^{(t-1)} + (1-\alpha) \x_i,\\
\label{eq:update-cluster_1}
\end{equation}
where $\bm{\tilde v}_i^{(t)}$ and $\bm{\hat v}_i^{(t)}$  unpdate in the same way, and $\alpha$ is set as 0.2 by default.

In implementation, we use DBSCAN algorithm~\cite{EsterDBSCAN:AAAI96} 
to generate the raw clusters. DBSCAN is a density-based clustering algorithm. It regards a data point as {density reachable} if the data point lies within a small distance threshold $d$ to other 
samples, where the parameter $d$ is the distance threshold to find neighboring point.

\begin{table*}[!t]
\centering
\setlength{\tabcolsep}{4.5mm}{
\begin{tabular}{lccccccc}
\hline		
\multirow{2}{*}{Dataset} & \multirow{2}{*}{Size} & \multicolumn{3}{c}{Subset} & \multirow{2}{*}{Identity} & \multirow{2}{*}{Camera} & \multirow{2}{*}{Clothes}\\
\cline{3-5}
& &Train&Query&Gallery& & & \\
\hline
DeepChange& 178, 407 & 75, 083 & 17, 527 &  62, 956 & 1, 121 & 17 & - \\
LTCC &  17, 119 & 9, 576 & 493 & 7, 050 & 152 & 12 & 14\\
PRCC & 33, 698 & 17, 896 & 3, 543 & 12, 259 & 221 & 3 & - \\
Celeb-ReID & 34, 186 & 20, 208 & 2, 972 & 11, 006 & 1, 052 & - & - \\
Celeb-ReID-Light & 10, 842 &  9, 021 & 887  &  934 &  590 & - & -\\
VC-Clothes& 19, 060 & 9, 449 & 1, 020 & 8, 591 & 256 & 4 & 3\\
\hline
\end{tabular}
}
\caption{Long-term Person Re-ID Datasets details.}
\label{tab:dataset}
\end{table*}

\myparagraph{Inference Procedure} 
After training, we keep only the ResNet $F(\cdot|\Theta)$ in for inference. 
We compute the distances between each image in the query and each image in the gallery using the feature obtained from the output of the first branch $F(\cdot |\Theta)$. We then  
sort the distances in ascending order to discover the matched results.

\begin{algorithm}[!t]
\caption{SiCL algorithm}
\label{alg:SiCL}
\textbf{Input}: Image $\II$ and Silhouette $\S$;\\
\textbf{Parameter}: Cluster distance $d$, Neighbor Searching Range $K$;\\
\textbf{Output}: $P_{best}$
\begin{algorithmic}[1] %[1] enables line numbers
\STATE Pre-train the two network branches on ImageNet;
\STATE Initialize the instance memory banks $\M$, $\tilde M$, $\hat M$ and set  $P = P_{best}=0$;
\WHILE{epoch $\leq$ total epoch}
\STATE Perform feature extraction to get $\X$, $\tilde \X$ and $\hat \X$;
\STATE Perform clustering on $\X$ to yield $m$ clusters $\C:=\{\C^{(1)}, \C^{(2)}, \cdots, \C^{(m)}\}$; 
\STATE Generate cluster-level neighbor set $\A$ through Eq~\ref{eq:cluster_sim};
\STATE Train SiCL with supervised infromation provided by $\A$ and $\C$, \ie, updating $\Theta$, $\Psi$, $\Omega$ and $\Theta^\prime$ via the total loss in Eq.~\ref{eq:AllLoss};
\STATE Update instance memory bank $\M$, $\tilde\M$ and $\hat M$ via Eq.~\ref{eq:update-cluster_1};
\STATE Evaluate the model performance $P$ with $F(\cdot|\Theta)$;
\IF{ $P > P_{best}$}
\STATE Output the best model $F(\cdot|\Theta)$ and set $P_{best} \leftarrow P$;
\ENDIF
\ENDWHILE
\STATE \textbf{return} $P_{best}$.
\end{algorithmic}
\end{algorithm}

\begin{table*}[!t]
\centering
\setlength{\tabcolsep}{3.3mm}{
\begin{tabular}{llcccccccccc}
\hline		
\multirow{3}{*}{Method} & \multirow{3}{*}{Reference} & \multicolumn{4}{c}{LTCC}&\multicolumn{6}{c}{PRCC}\\
\cline{3-12}
& &\multicolumn{2}{c}{C-C}&\multicolumn{2}{c}{General}&\multicolumn{3}{c}{C-C}&\multicolumn{3}{c}{General}\\
\cline{3-12}  
& & mAP & R-1 &mAP & R-1 &mAP & R-1 & R-10 &mAP & R-1 & R-10\\

\hline
\multicolumn{4}{l}{Supervised Method(ST-ReID)} \\
\hline

HACNN & CVPR'18 & 9.30 & 21.6 & 26.7 & 60.2 & -  & 21.8 & 59.4 & - & 82.5 & 98.1\\
PCB & ECCV'18 & 10.0 & 23.5 & 30.6 & 38.7 & 38.7 & 22.8 & 61.4  & 97.0 & 86.8 & 99.8\\
\hline
\multicolumn{4}{l}{Supervised Method(LT-ReID)} \\
\hline
CESD & ACCV'20 & 12.4 & 26.2 & 34.3 & 71.4 &  - & - & - &  - & - & -\\
RGA-SC& CVPR'20 & 14.0 & 31.4 & 27.5  & 65.0 & - & 42.3 & 79.4 & - & 98.4 & 100\\
IANet & CVPR'19 & 12.6 & 25.0 & 31.0 & 63.7& 45.9 & 46.3 & - & 98.3 & 99.4 & - \\
GI-ReID & CVPR'22 & 10.4 & 23.7 & 29.4 & 63.2 &  - & - & - &  - & - & - \\
RCSANet & CVPR'21 & - & - & - & - & 48.6 & 50.2 & - & 97.2 & 100 & -  \\
3DSL & CVPR'21 & 14.8 & 31.2 & - & - & - & 51.3 &-  & - & - & -\\
FSAM & CVPR'21 & 16.2 & 28.5 & 25.4 & 73.2 & - & 54.5 & 86.4 & -  & 98.8 & 100\\
CAL & CVPR'22 & 18.0 & 40.1 & 40.8 & 74.2 & 55.8 & 55.2 & - & 99.8 & 100 & -  \\ 
CCAT & IJCNN'22 & 19.5 & 29.1 & 50.2 & 87.2 & - &69.7 & 89.0 & - & 96.2 & 100\\ 
\hline
\multicolumn{4}{l}{Unsupervised Method} \\
\hline
% \hline
SpCL & NeurIPS'20 & 7.60 & 15.3  & 21.2 & 47.3 & 45.2 & 33.2 & 71.3 & 90.6 & 86.4 & 98.3 \\
C3AB & PR'22 & 8.30 & 15.2 & 20.7 & 46.7 & 48.6 & 36.7 & 74.0 & 90.2 & 88.3 & 98.1 \\
CACL & TIP'22 & 6.20 & 9.80 & 22.3 & 45.6 & 52.1 & 41.7 & 79.8 & 94.7 & 90.9 & 99.9\\
CC& ACCV'22 & 6.00 & 7.40 & 11.0 & 17.0 & 46.3 & 34.4 & 74.4  & 94.4 & 90.2 & 99.9 \\
ICE& ICCV'21 & 7.10 & 14.5 & 28.4 & 61.1 & 48.0 & 34.8 & 74.2 & 95.9 & 93.6 & 99.9 \\
ICE* & ICCV'21 & 10.1 & 16.3  & 22.6 & 44.0 & 45.5 & 32.6 & 72.3 &95.7 & 93.3  & 99.8    \\
PPLR & CVPR'22 & 4.40 & 4.80 & 6.00 & 11.2  & 51.4 & 40.0 & 75.2  & 91.7 & 87.4 & 99.8  \\
\hline
\bf{SiCL} & \bf{Ours}  & \bf{10.1} & \bf{20.7} & \bf{27.6} & \bf{57.6} &  \bf{55.4} & \bf{43.2} & \bf{80.2}  &  \bf{96.2} & \bf{93.8} & \bf{99.4}   \\
\hline
\end{tabular}
}
\caption{Comparison with the state-of-the-art methods on LTCC and PRCC, `C-C' means clothes change setting, `General' means general setting. 
{ `*'  means using the camera label as side-information.
}}
\label{tab:SOTALTCC}
\end{table*}

\begin{table}[tb]
\centering
\setlength{\tabcolsep}{2.3mm}{
\begin{tabular}{llcccc}
\hline		
\multirow{3}{*}{Method} & \multirow{3}{*}{Reference} & \multicolumn{4}{c}{VC-Clothes}\\
\cline{3-6}
& &\multicolumn{2}{l}{C-C}&\multicolumn{2}{l}{General}\\
\cline{3-6}  
& & mAP & R-1 & mAP & R-1\\
\hline
\multicolumn{4}{l}{Supervised Method(ST-ReID)} \\
\hline
PCB & ECCV'18 &  30.9 & 34.5 & 83.3 & 86.2 \\
\hline
\multicolumn{4}{l}{Supervised Method(LT-ReID)} \\
\hline
HACNN & CVPR'18 & 62.2 & 62.0 & 94.3 & 94.7\\
RGA-SC& CVPR'20 & 67.4 & 71.1 & 94.8 & 95.4\\
FSAM & CVPR'21 & 78.9 & 78.6 & 94.8 & 94.7 \\
CCAT & IJCNN'22 & 76.8 & 83.5 & 85.5 & 92.7 \\
\hline
\multicolumn{4}{l}{Unsupervised Method} \\
\hline
SpCL & NeurIPS'20 & 38.2 & 46.2 & 61.0 & 77.8 \\
C3AB & PR'22 & 44.1 & 52.0 & 65.0 & 81.0\\
CACl & TIP'22  & 49.7 & 58.9 & 68.0 &82.4\\
CC & ACCV'22 & 25.7 & 31.0 & 45.1 & 62.4 \\
ICE & ICCV'21& 28.7 & 34.5 & 51.2 & 69.3 \\
ICE* & ICCV'21 & 28.5 & 31.4 & 51.8 & 70.1 \\
PPLR & CVPR'22 & 23.1 & 32.5 & 47.7 & 68.1\\

\hline
\bf{SiCL} & \bf{Ours}  & \bf{63.9} & \bf{71.7} & \bf{77.2} & \bf{87.9} \\
\hline
\end{tabular}
}
\caption{Comparison on VC-Clothes.
}
\label{tab:SOTAVC}
\end{table}

\begin{table}[!t]
\centering
\setlength{\tabcolsep}{2.1mm}{
\begin{tabular}{llcccc}
\hline		
\multirow{2}{*}{Method} & \multirow{2}{*}{Reference} & \multicolumn{2}{c}{Celeb-ReID}&\multicolumn{2}{c}{Celeb-Light}\\
\cline{3-6}  
& & mAP & R-1 & mAP & R-1\\
\hline
\multicolumn{4}{l}{Supervised Method(ST-ReID)} \\
\hline
TS & TOMM'17 & 7.80  & 36.3 & - & - \\
\hline
\multicolumn{4}{l}{Supervised Method(LT-ReID)} \\
\hline
MLFN & CVPR'18 & 6.00 & 41.4 & 6.30 & 10.6 \\
HACNN & CVPR'18 & 9.50 & 47.6 & 11.5 & 16.2\\ 
PA& ECCV'18 & 6.40 & 19.4 & - & - \\
PCB & ECCV'18 & 8.20 & 37.1 & - & - \\
MGN & MM'18 & 10.8 & 49.0 & 13.9 & 21.5 \\
DG-Net& CVPR'19 & 10.6 & 50.1 & 12.6 & 23.5\\
Celeb & IJCNN'19 & - & - & 14.0 & 26.8 \\
Re-IDCaps & TCSVT'19 & 9.80 & 51.2 & 11.2 & 20.3 \\
RCSANet & CVPR'21 & 11.9 & 55.6 & 16.7 & 29.5 \\
\hline
% \hline
\multicolumn{4}{l}{Unsupervised Method} \\
\hline
% \hline
SpCL & NeurIPS'20 & 4.60 & 39.6 & 3.60 & 5.30 \\
C3AB& PR'22 & 4.80 & 41.0 & 3.70 & 5.10 \\
CACL & TIP'22  & 5.10 & 42.3 & 3.60  & 4.30 \\
CC & ACCV'22 & 3.40 & 32.8 & 3.20 & 4.40 \\
ICE& ICCV'21 & 4.90 & 40.7 & 5.00 & 7.10 \\
PPLR & CVPR'22 & 4.80 & 41.3 &  4.30 & 6.20\\
\hline
\bf{SiCL} & \bf{Ours}  & \bf{5.60} & \bf{45.4} &  \bf{5.60} & \bf{10.5} \\
\hline
\end{tabular}
}
\caption{Comparison with the state-of-the-art methods on Celeb-ReID and Celeb-ReID-Light.}
\label{tab:SOTACele}
\end{table}

%-------------------------------------------------------------------------

\section{Experiments}
\label{sec:Experiments}

\subsection{Experimental Setting}

\myparagraph{Datasets}
\label{p:dataset-and-preprocessing}
We evaluate SiCL on {six} clothes change re-id datasets: 
LTCC~\cite{Change:1}, 
PRCC~\cite{yang2019person}, 
VC-Clothes~\cite{vcclothes}, 
Celeb-ReID~\cite{celebreid}, 
Celeb-ReID-Light~\cite{celebreidlight} and DeepChange~\cite{Deepchange}.
Table~\ref{tab:dataset} shows an overview of dataset in details.

\myparagraph{Protocols and Metrics}
\label{p:protocols-and-metrics}
Different from the traditional short-term person \reid{} setting, 
there are two evaluation protocols for long-term person \reid{}: a) general setting and b) clothes-change setting. Specifically, for a query image, the general setting is looking for cross-camera matching samples of the same identity, while the clothes-change setting additionally demands the same identity with inconsistent clothing.
For LTCC~\cite{Change:1} and PRCC~\cite{yang2019person}, we report performance with both clothes-change setting  and general setting. As for Celeb-ReID~\cite{celebreid}, Celeb-ReID-Light~\cite{celebreidlight} DeepChange~\cite{Deepchange} and VC-Clothes~\cite{vcclothes}, we report the general setting performance.
We use both Cumulated Matching Characteristics (CMC) and mean average precision (mAP) as retrieval accuracy metrics.

\myparagraph{Implementation Details} In SiCL, we use ResNet-50~\cite{he:CVPR16resnet} pre-trained on ImageNet~\cite{Krizhevsky:NIPS12} for both  network branches.
The features dimension $D = 2048$.
We use the output $\x_i$ of the first branch $F(\cdot|\Theta)$ to perform clustering, where $\x_i \in  \RR^D$. 
The prediction layer $G(\cdot|\Psi)$ is a $D \times D$ full connection layer, the fusion layer $R(\cdot|\Omega)$ is a $2D \times D$ full connection layer. The cluster distance $d = 0.6$ for PRCC, LTCC, Celeb-ReID and Celeb-ReID-Light datasets, and $d = 0.7$ for VC-Clothes.
We optimize the network through Adam optimizer \cite{Kingma:arXiv2014} with a weight decay of 0.0005 and train the network with 60 epochs in total. The learning rate is initially set as 0.00035 and decreased to one-tenth per 20 epochs. 
The batch size is set to 64. 
The temperature coefficient $\tau$ in Eq.~\ref{eq:softmax-cal} is set to $0.05$ and the update %upgrade 
factor $\alpha$ in Eq.~\ref{eq:update-cluster_1} is set to $0.2$. The $K$ in Eq.~\ref{eq:Curriculumneighbor} is set as 10 on LTCC, Celeb-ReID
and Celeb-ReID-Light, 3 on VC-Clothes and 5 on PRCC, the effect of using a different value of $K$ will be test later.\footnote{The code for the work will be released upon the acceptance of the paper.}

%-------------------------------------------------------------------------

\subsection{Comparison to State-of-the-art Methods} 
\label{subsec:comp-to-sota}

\myparagraph{Competitors}
\label{p:baselines}
To construct a preliminary benchmark and conduct a thorough comparison, we evaluate some \sota{} of short-term unsupervised \reid{} models, which are able to achieve competitive performance under an unsupervised short-term setting, including SpCL~\cite{Ge:NIPS20}, CC~\cite{tanping:arxiv2020}, CACL~\cite{TIPlmk}, C3AB~\cite{lmkpr}, ICE~\cite{ICE}, PPLR~\cite{PPLR}.
We retrain and evaluate these unsupervised methods on long-term datasets, including LTCC, PRCC, Celeb-ReID, Celeb-ReID-Light and VC-Clothes. 
At the same time, we also compare with other supervised long-term person \reid{} methods such as HACNN~\cite{li:HACNNCVPR18}, PCB~\cite{sun:PCBECCV18}, CESD~\cite{qian:CESDACCV20}, RGA-SC~\cite{zhang:RGASCCVPR20}, IANet~\cite{hou:IANETCVPR19}, GI-ReID~\cite{jin:GI-ReIDCVPR22}, RCSANet~\cite{huang:RCSANetCVPR21}, 3DSL~\cite{chen:3DSLCVPR21}, FSAM~\cite{hong:FSAMCVPR21}, CAL~\cite{CCgu:CVPR22}, CCAT~\cite{ren:CCATIJCNN22}. 
The comparison results of the state-of-the-art unsupervised short-term  person \reid{} methods and supervised methods are shown in Tables~\ref{tab:SOTALTCC},~\ref{tab:SOTACele} and~\ref{tab:SOTAVC}.
From the results in these tables we can read that our SiCL outperforms much better than all unsupervised short-term methods and even comparable to some supervised long-term methods.

Particularly, we can observe that 
SiCL yields much higher performance than the short-term unsupervised person re-id methods in clothes-change setting. Yet, the differences are reduced and even vanished in general settings (\eg, ICE results on LTCC).  This further demonstrates the dependency of short-term re-id methods on clothes features for person matching. 
In addition, we discovered that SiCL performs poorly on certain datasets, such as Celeb-ReID-Light; we will investigate the underlying causes in \textit{Supplementary Material}.

\subsection{Ablation Study}
In this section, we conduct a series of ablation experiments to evaluating each component in SiCL architecture, \ie, contrastive learning framework, the neighbor contrastive learning module $\mathcal{L}_{N}$, separately.
In addition, we substitute the feature fusion operation with different branch feature. 

\myparagraph{ Baseline Setting} In the baseline network, we employ the prototypical contrastive learning module and the cross-view contrastive learning module and train the model with the corresponding loss functions. The training procedure and memory updating strategy are kept the same as SiCL.

The ablation study results are presented Table~\ref{tab:ablationstudy}. We can read that when each component is added separately, the performance increases. This verifies that each component contributes to improved performance.

\myparagraph{Other Experiments}
 We have provided more comprehensive and complex results of the ablation experiments in the \textit{Supplementary Material}, including the range of neighbour search $K$, the clustering distance parameter $d$, and others.

%-------------------------------------------------------------------------

%-------------------------------------------------------------------------

\begin{table}[!t]
\centering
\setlength{\tabcolsep}{2.2mm}{
\begin{tabular}{lcccccc}
\hline		
\multirow{2}{*}{Method}  &\multicolumn{2}{c}{VC}&\multicolumn{2}{c}{PRCC}&\multicolumn{2}{c}{LTCC}\\
\cline{2-7}  
&mAP & R-1 &mAP & R-1 &mAP & R-1  \\
\hline
{C-C} \\
\hline
{Single*}    &   44.4 & 54.0   & 48.4 & 37.7  &  9.20 & 17.1   \\
{Baseline}   &  62.8 & 71.4   &  54.2& 42.5 & 9.80 & 18.6 \\
{SiCL}    & 63.9 & 71.7  & 55.4 &   43.2  & 10.1 & 20.7  \\
\hline
\end{tabular}
}
\caption{Ablation Study on LTCC, PRCC and VC-Clothes, Single* means only use RGB images in training without contrastive learning.}
\label{tab:ablationstudy}
\end{table}

\subsection{More Evaluations and Analysis}

\myparagraph{Different Operations for Cluster-level Neighbor Searching}
To determine the requirement of feature fusion, we prepare a set of tests employing different operations, \ie  using $\x$ and $\bm{\tilde x}$, to find the cluster neighbors.
The experimental results are listed  in Table~\ref{tab:featurefusion}. We can read that only using $\bm{\tilde x}$ in the cluster neighbor searching stage will also yield acceptable performance in finding cluster neighbors. This observation confirms that rich information on the high-level neighbor structure may be gleaned using just a silhouette source. However, the results obtained from the LTCC dataset indicate that the feature fusion operations did not yield optimal performance. The rationales behind this observation will be elaborated upon in the supplementary materials.

\begin{table}[!t]
\centering
\setlength{\tabcolsep}{1.9mm}{
\begin{tabular}{lcccccc}
\hline		
\multirow{2}{*}{Operation} &\multicolumn{2}{c}{VC}&\multicolumn{2}{c}{PRCC}&\multicolumn{2}{c}{LTCC}\\
\cline{2-7}  
& mAP & R-1 &mAP & R-1 &mAP & R-1  \\
\hline
\multicolumn{1}{l}{C-C} \\
\hline
{ $\x$} & 61.8&69.4 & 55.1&42.9 & 10.4&21.4  \\
{ $\bm{\tilde x}$} & 63.6&71.4  & 53.8&42.4 & 10.9&21.9  \\{ SiCL} & 63.9 & 71.7  & 55.4 &   43.2  & 10.1 & 20.7  \\
\hline
\end{tabular}
}
\caption{Experiments Results of Various Cluster Neighbor Searching Operations on LTCC, PRCC, and VC-Clothes.}
\label{tab:featurefusion}
\end{table}

\myparagraph{Performance Evaluation on DeepChange} 
DeepChange is the latest, largest, and realistic person \reid{} benchmark with clothes change. We conduct experiments to evaluate the performance of SiCL and SpCL on this dataset, and report the results in Table~\ref{tab:Deepchange-SiCL}. 
Moreover, we also listed the results of supervised training methods using the different backbones provided by DeepChange~\cite{Deepchange}.

\begin{table}[!t]
\centering
    \setlength{\tabcolsep}{3.1mm}{
    \begin{tabular}{llccc}
    \hline		
     \multirow{2}{*}{Method}& \multirow{2}{*}{Backbone}&\multicolumn{3}{c}{DeepChange}\\
     \cline{3-5}  
    &  & mAP & R-1 &R-10 \\
    \hline
    \multicolumn{2}{l}{Supervised Method} \\
    \hline
    DeepChange &{ResNet-50} & 9.62 &  36.6 & 55.5     \\
    DeepChange &ReIDCaps  & 13.2 & 44.2 & 62.0 \\
    DeepChange &ViT B16 & 14.9 & 47.9 & 67.3 \\
    \hline
    \multicolumn{2}{l}{Unsupervised Method} \\
    \hline
    SpCL& ResNet-50 & 8.90 & 32.9 &47.2 \\
    {SiCL}& ResNet-50 & 12.4 & 41.0  & 56.0 \\
    \hline
    \end{tabular}
    }
    \caption{Experimental Results on DeepChange.}
    \label{tab:Deepchange-SiCL}
    \end{table}

\section{Conclusion}

We have addressed a challenging task: unsupervised long-term person \reid{} with clothes change. Specifically, we proposed a silhouette-driven contrastive learning approach, termed SiCL, which takes the silhouette masks into contrastive learning and constructs a hierarchically neighborhood structure to %drive 
facilitate the training. 
By leveraging the silhouette feature and the hierarchical neighborhood relationship, SiCL is able to effectively exploit the invariance within and between RGB images and silhouette masks to learn more effective cross-clothes features. 
We conducted extensive experiments on six widely-used
re-id datasets with clothes change. Experimental results demonstrated the superiority of our proposed approach. To the best of our knowledge, this is the first time that the unsupervised long-term person \reid{} problem has been tackled. In addition, our systematic evaluations can serve as a benchmark for the {unsupervised} long-term person re-id community.
{
    \small
    \bibliographystyle{ieeenat_fullname}
    \bibliography{main}
}

\clearpage
\setcounter{page}{1}
\maketitlesupplementary

In the supporting materials, we provide the implementation details of the experiments, such as evaluations on using different $K$, $d$, and a set of visualization results. Besides, we also discuss the limitation and the future work. 

\section{Implementation Details and Supplementary Experiments}

\subsection{Implementation Details}
Our experiments are implemented with PyTorch and run on a server equipped with an Intel(R) Xeon(R) CPU E5-2630-v4 and four Nvidia 3090 GPUs of memory 24 GB.
The contrastive learning framework is based on CACL.

\subsection{More Evaluation and Analysis}

\myparagraph{Ablation Study on Celeb-ReID and Celeb-ReID-Light }
The ablation study results are presented in Table~\ref{tab:ablationstudyceleb}. We can read that when each component is added separately, the performance increases. This verifies that each component contributes to improved performance.

\myparagraph{Performance on Celeb-ReID-Light} 
For the LTCC, PRCC, and VC-Clothes datasets, SiCL does not vary substantially from the supervised methods; however, the difference is apparent on the Celeb-ReID-Light dataset.
Due to the small size of the Celeb-ReID-Light dataset, the look of the same person apparel differs substantially while including just one picture per clothing type. 
This is a significant difficulty for the clustering-based methods, as it is difficult for the model to accurately recognize the same pedestrians wearing various garments during training.

\myparagraph{Evaluation on Parameters in DBSCAN} 
We conduct experiments on LTCC and VC-Clothes to evaluate the effects of changing the  parameter $d$.
Experiments are recorded in Table~\ref{Tab:cluster-parameter}. we can find
that while $d = 0.6$, SiCL demonstrates the highest performance on LTCC. 
We can observe that  SiCL is comparatively sensitive to $d$. 
This suggests that SiCL is somewhat reliant on the quality of the clustering. This is because if the clustering quality is low, the instance neighbor information will also be in error. Such errors can cumulatively at the clustering level, which negatively impacts the model training.
Therefore, one of our forthcoming research directions is to make SiCL less sensitive to the clustering parameter in the future.

\begin{table}[!t]
\centering
\caption{Performance Comparison of different cluster parameter $d$ (the maximum  distance  between  neighbor points) on SiCL.}
\setlength{\tabcolsep}{4.5mm}{
\begin{tabular}{lcccc}
\hline		
\multirow{2}{*}{Eps}  & \multicolumn{2}{c}{VC-Clothes} & \multicolumn{2}{c}{LTCC}\\
\cline{2-5}
&  mAP & R-1 & mAP & R-1\\
\hline
{General} & \\
\hline
 0.8 &8.70&15.7 & 4.30 & 9.90 \\ 
 0.7 &77.2&87.9 & 11.8& 25.2\\ 
 0.6 &76.5&87.4 & 27.6 & 57.6 \\
 0.5 &70.4&82.8 & 22.9 & 46.5\\
 0.4 &64.3&78.0 & 18.7 & 43.6 \\

\hline
\end{tabular}
\label{Tab:cluster-parameter}
}
\end{table}

\myparagraph{Different operations for cluster-level neighbor searching}
To determine the requirement of feature fusion, we prepare 
a set of tests employing different operations, such as using $\x$, $\bm{\tilde x}$, to find the cluster-level neighbors.
The experimental results 
are listed in Table~\ref{tab:featurefusion}. 
We can find that using only $\bm{\tilde x}$ during the cluster-level neighbour searching stage also yield satisfactory results. Conversely, using only $\x$ to search the cluster-level neighbor would have a detrimental effect on the model performance on both VC-Clothes and PRCC.
This observation confirms that rich clothes-independent information can be gleaned by just using the silhouette feature.

\begin{table}[!t]
\centering
\setlength{\tabcolsep}{4mm}{
\begin{tabular}{lcccc}
\hline		
\multirow{2}{*}{Method}  &\multicolumn{2}{c}{Celeb-ReID}&\multicolumn{2}{c}{Celeb-ReID-Light}\\
\cline{2-5}  
&mAP & R-1 &mAP & R-1\\
\hline
{C-C} \\
\hline
{Single*}   &4.80& 40.5&5.00& 8.50 \\ 
{Baseline}   &  5.60&44.2&5.20&10.0 \\
{SiCL}    &6.00&45.0&5.40&9.50\\ 
\hline
\end{tabular}
}
\caption{Ablation Study on Celeb-ReID and Celeb-ReID-Light, Single* means only use RGB images in training without contrastive learning.}
\label{tab:ablationstudyceleb}
\end{table}

\begin{table}[!t]
\centering
\setlength{\tabcolsep}{2mm}{
\begin{tabular}{lcccccc}
\hline		
\multirow{2}{*}{Operation} &\multicolumn{2}{c}{VC}&\multicolumn{2}{c}{PRCC}&\multicolumn{2}{c}{LTCC}\\
\cline{2-7}  
& mAP & R-1 &mAP & R-1 &mAP & R-1  \\
\hline
\multicolumn{1}{c}{C-C} \\
\hline
{Baseline} & 62.8 & 71.4 & 54.2 & 42.5 & 9.80 & 18.6 \\
{ $\x$}  & 61.8 & 69.4  & 54.1 &  42.4 & 10.4 & 20.4  \\
{ $\bm{\tilde x}$} & 63.6&71.4 & 54.6 &42.6 & 10.5&20.9  \\
{ SiCL} & 63.9 & 71.7  & 55.4&   43.2  & 10.1 & 20.7  \\
\hline
\end{tabular}
}
\caption{Experiments Results of Various Cluster-Level Neighbor Searching Operations on LTCC, PRCC, and VC-Clothes.}
\label{tab:featurefusion}
\end{table}

\myparagraph{Performance with different $K$}
Table~\ref{tab:kneighbor-SiCL} illustrates the effect of varying 
cluster-level neighbor search parameters $K$ across various datasets. 
We can find that our SiCL is not sensitive to the change  
of the cluster-level neighbor search 
parameter $K$ on PRCC. However, for the VC-Clothes dataset, when the neighbour search range is less than 6, the cluster-level neighbour information seems to play a negative role and does not bring any performance improvement to the model. We believe that this problem might be caused due to low accuracy of cluster-level neighbor selection when the value of $K$ is small. Hence, our future work will investigate how to enhance the quality of cluster-level neighbor selection.

\begin{table}[!t]
\centering
    \setlength{\tabcolsep}{2.8mm}{
    \begin{tabular}{lcccccc}
    \hline		
    \multirow{3}{*}{$K$} &\multicolumn{4}{c}{VC-Clothes}&\multicolumn{2}{c}{PRCC}\\
    \cline{2-7}  
    & \multicolumn{2}{c}{C-C} &  \multicolumn{2}{c}{General} & \multicolumn{2}{c}{C-C}  \\
    \cline{2-7}  
    & mAP & R-1 &mAP & R-1 &mAP & R-1  \\
    \hline
     10 &63.9&71.7 & 77.2&87.9 & 54.4& 42.6  \\ 
     8 & 63.4 & 71.2 &76.8&87.7 &55.2& 43.8 \\ 
     6 & 62.5 & 70.0  &75.5&86.9& 54.5 & 42.5 \\ 
     4 & 63.0 & 70.5 &77.2&87.1& 54.1 & 43.1 \\ 
     2 & 63.1 & 69.8  &74.5&86.1 & 54.6 & 42.8 \\
    \hline
    \end{tabular}
    }
     \caption{Illustration for the model performance with different $K$ on PRCC and VC-Clothes.}
     \label{tab:kneighbor-SiCL}
    \end{table}

\begin{figure*}
    \centering
    \begin{subfigure}[SiCL vs. CACL on LTCC]{0.49\textwidth}
        \includegraphics[scale=0.25]{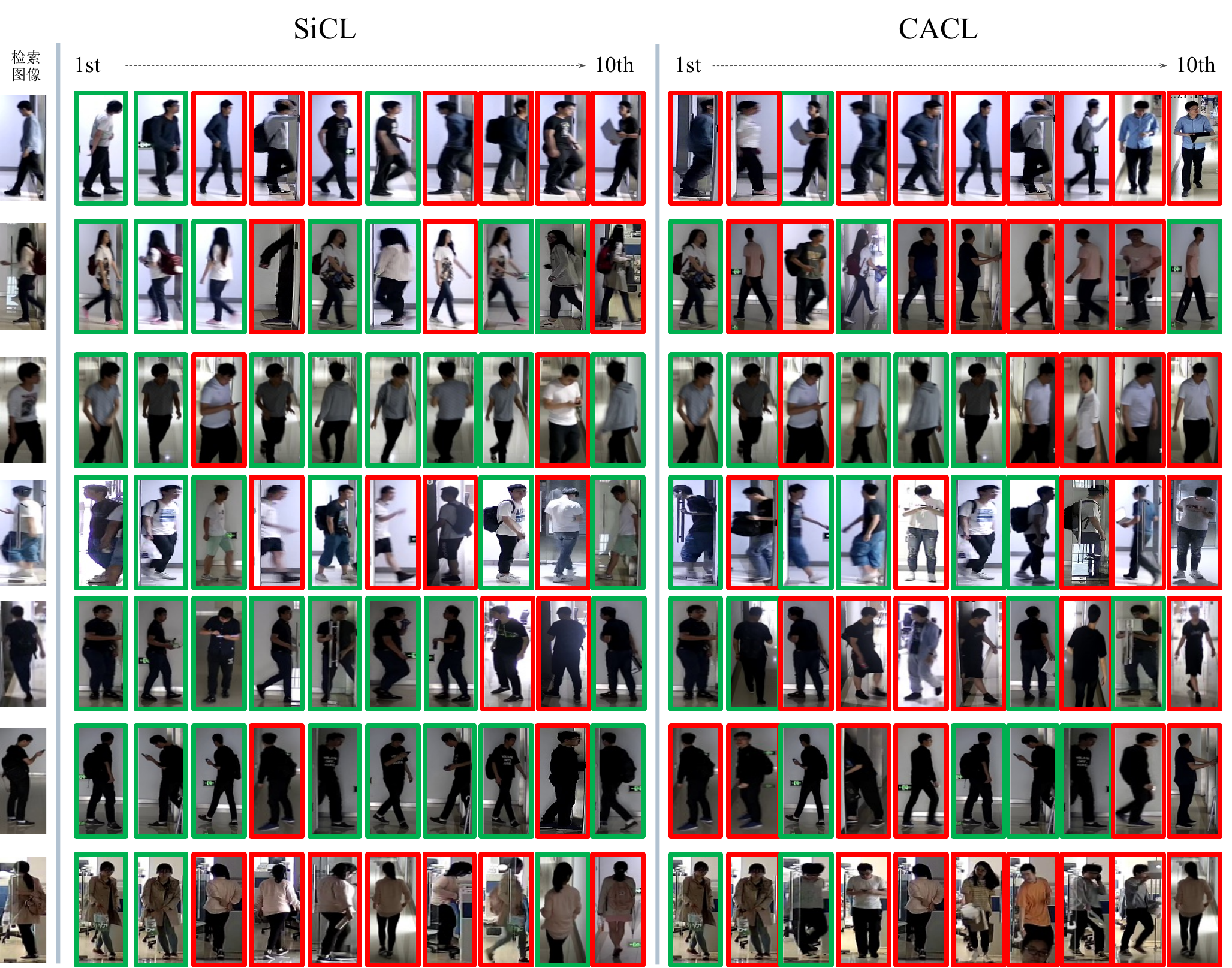}
    \end{subfigure}
    \begin{subfigure}[SiCL vs. CACL on VC-Cltohess]{0.49\textwidth}
        \includegraphics[scale=0.25]{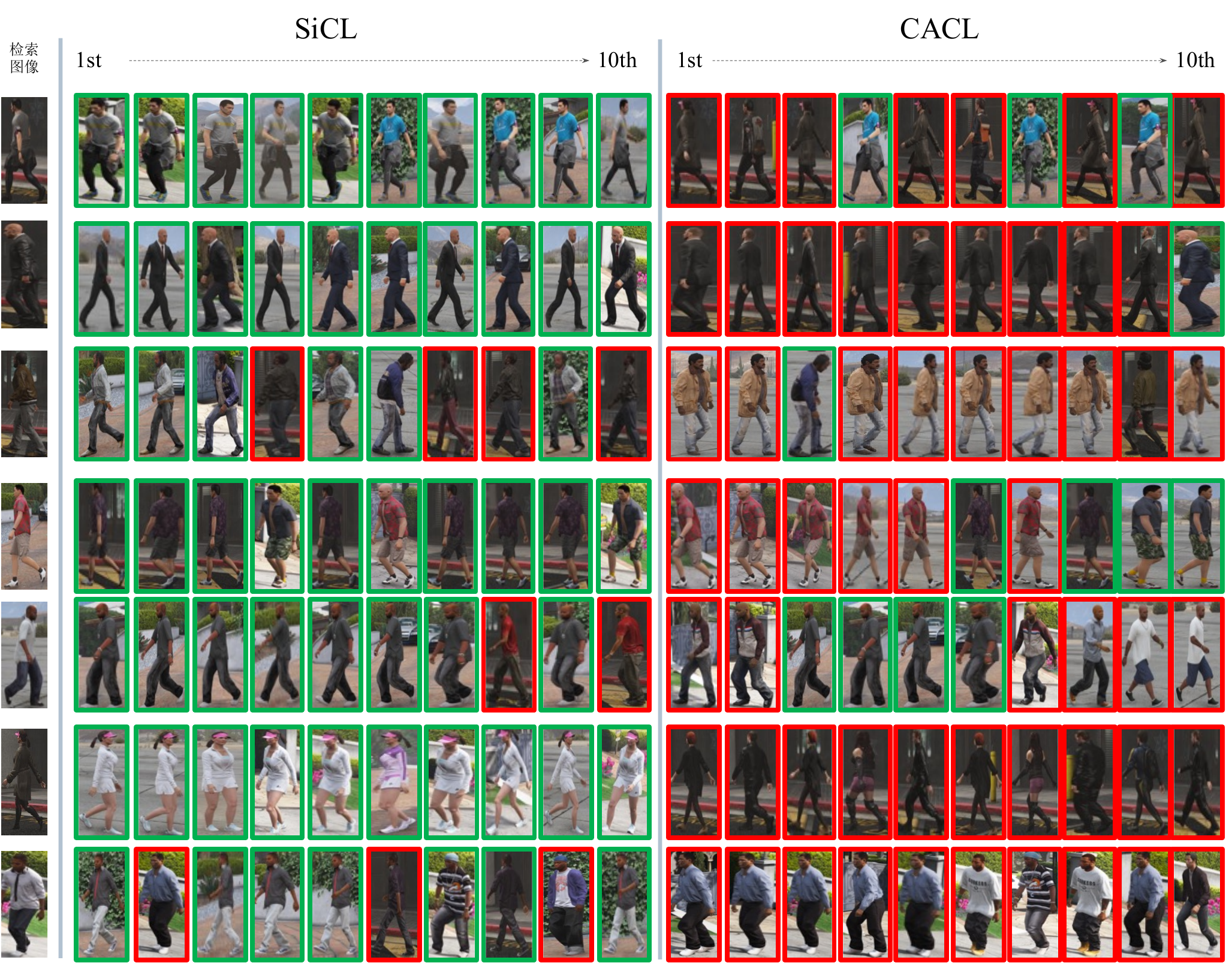}
    \end{subfigure}
    \caption{   Visualization of the top-10 best matched images on LTCC and VC-Clothes. We show the top-10 best matching samples in the gallery set for the query sample with CACL and our proposed SiCL. The images with frames in green and in red are the correctly matched images and mismatched images, respectively.}
    \label{fig:vision_SiCL_5}
\end{figure*}
        
\begin{figure*}
    \centering
    \begin{subfigure}[b]{0.49\textwidth}
        \includegraphics[scale=0.25]{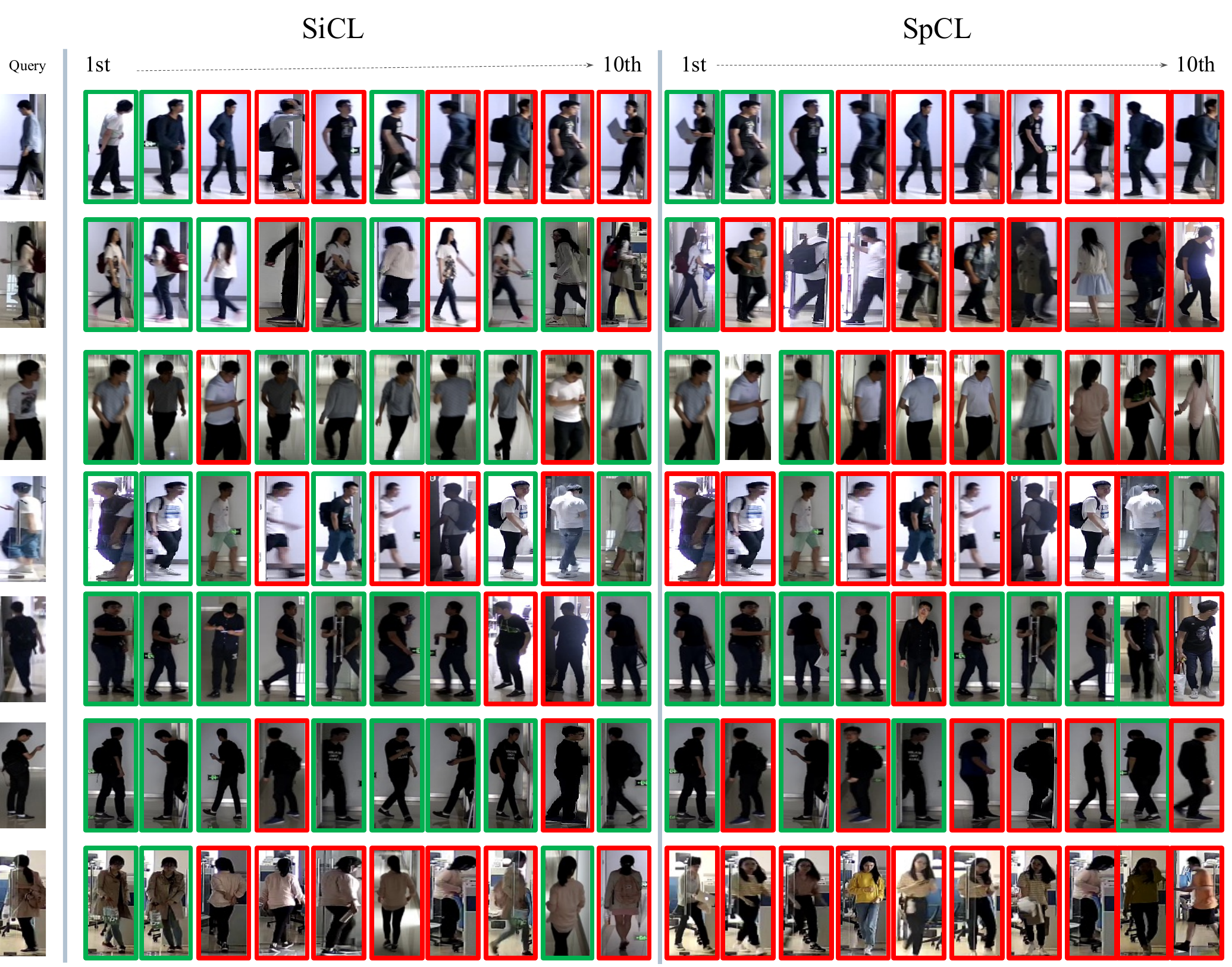}
    \end{subfigure}
    \begin{subfigure}[b]{0.49\textwidth}
        \includegraphics[scale=0.25]{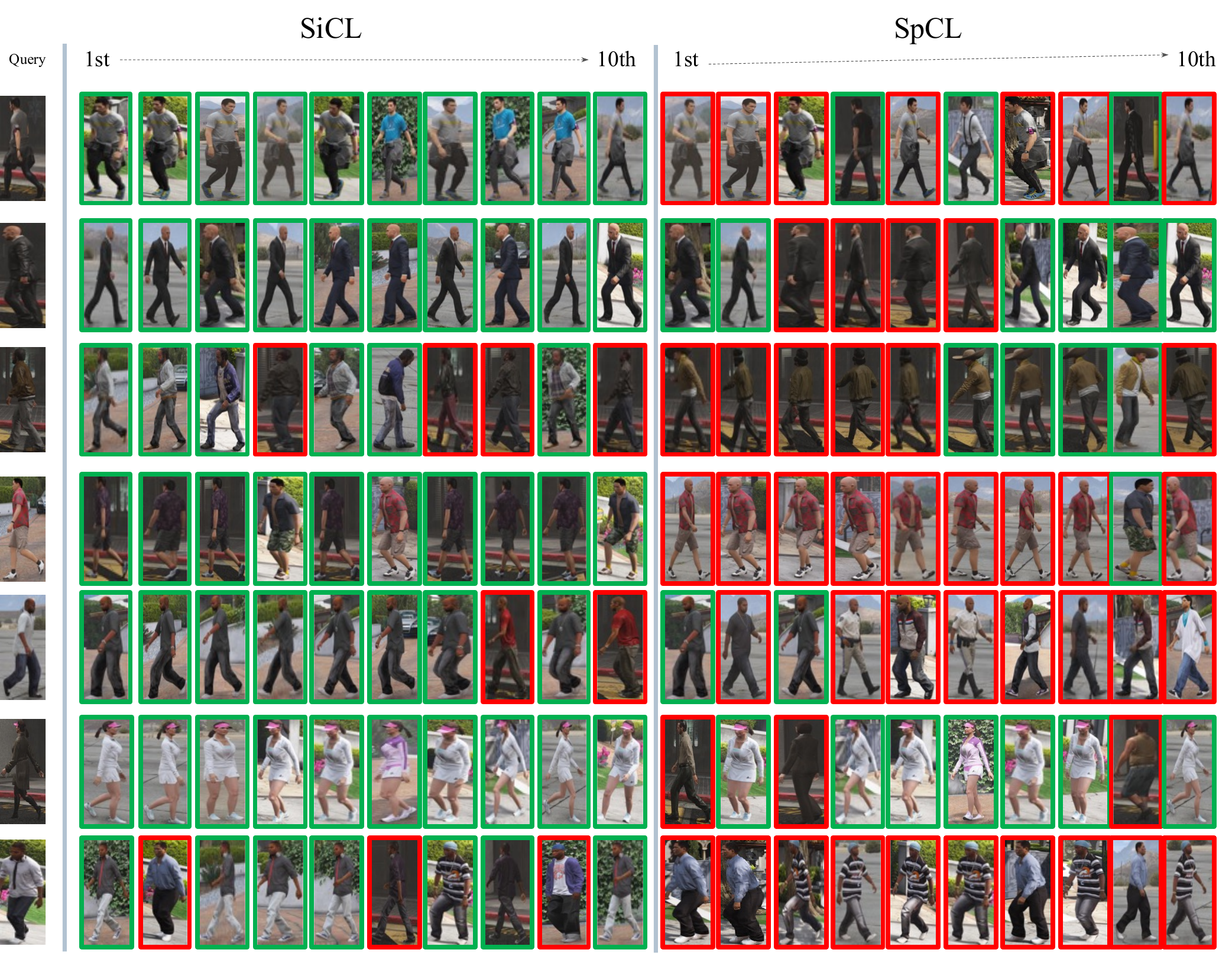}
    \end{subfigure}
    \caption{  Visualization of the top-10 best matched images on LTCC and VC-Clothes. We show the top-10 best matching samples in the gallery set for the query sample with SpCL and our proposed SiCL. The images with frames in green and in red are the correctly matched images and mismatched images, respectively.}
    \label{fig:vision_SiCL_4}
\end{figure*}

\myparagraph{Visualization}
To gain some intuitive understanding of the performance of our proposed SiCL, we conduct a set of data visualization experiments on LTCC and VC-Clothes to visualize selected query samples with the top-10 best matching images in the gallery set, and display the visualization results in Figure~\ref{fig:vision_SiCL_5} and Figure~\ref{fig:vision_SiCL_4}.
Compared to CACL and SpCL, SiCL yields more precise matching results. The majority of the incorrect samples matched by other methods are in the same color as the query sample. These findings imply that SiCL can successfully learn more clothing invariance features and hence identify more precise matches.

\section{Limitation and Future Work}
In experiments,  we construct a hierarchical neighbor structure using person silhouette masks and person images to drive the model learning in cross-clothes invariance.  Experiments have shown that clothe-independent features can provide the correct guidance for the model. Therefore, it is promising to research generating higher quality person silhouette masks without supervision to further enrich the representation capacity, improve the stability and enhance the overall performance of the proposed framework.

Besides, pedestrian silhouette, person clothes-independent semantic sources comprise numerous categories, such as gait, skeleton. As the future work, it is interesting to investigate combine multiple types of sources to enhance the model performance.

% \bibliography{aaai24}

\end{document}